\documentclass{article}
\usepackage{cite}
\usepackage{color}
\usepackage{amsmath,amssymb,amsfonts}
\usepackage{algorithmic}
\usepackage{graphicx}
\usepackage{textcomp}
\usepackage[margin=1in]{geometry}
\usepackage{xcolor}
\usepackage[hidelinks]{hyperref}
\renewcommand{\epsilon}{\varepsilon}
\newcommand{\worldone}{\texttt{Transfermania}}
\newcommand{\worldtwo}{\texttt{Privacyland}}

\newcommand{\citet}{\cite}
\newcommand{\citep}{\cite}

\begin{document}

\title{Position: Considerations for Differentially Private Learning with Large-Scale Public Pretraining\footnote{Authors are listed in reverse alphabetical order.}}
%\title{Private Machine Learning in the Era of Massive Public Datasets}
%\author{Gautam Kamath \and Florian Tram\`er}
\author{Florian Tram\`er\thanks{ETH Zurich. \texttt{florian.tramer@inf.ethz.ch}.} \and Gautam Kamath\thanks{Cheriton School of Computer Science, University of Waterloo and Vector Institute. {\tt g@csail.mit.edu}.} \and Nicholas Carlini\thanks{Google DeepMind. \texttt{nicholas@carlini.com}.}}
\date{}

\maketitle

\begin{abstract}
The performance of differentially private machine learning can be boosted significantly by leveraging the transfer learning capabilities of non-private models pretrained on large \emph{public} datasets.
We critically review this approach.

We primarily question whether the use of large Web-scraped datasets \emph{should} be viewed as differential-privacy-preserving. We caution that publicizing these models pretrained on Web data as ``private'' could lead to harm and erode the public's trust in differential privacy as a meaningful definition of privacy.

Beyond the \emph{privacy} considerations of using public data, we further question the \emph{utility} of this paradigm. We scrutinize whether existing machine learning benchmarks are appropriate for measuring the ability of pretrained models to generalize to sensitive domains, which may be poorly represented in public Web data.
Finally, we notice that pretraining has been especially impactful for the largest available models---models sufficiently large to prohibit end users running them on their own devices. Thus, deploying such models today could be a net loss for privacy, as it would require (private) data to be outsourced to a more compute-powerful third party.

We conclude by discussing potential paths forward for the field of private learning, as public pretraining becomes more popular and powerful.

\end{abstract}
\section{Introduction}

While machine learning models have made tremendous progress at learning \emph{generalizable concepts} from data at scale, these models also frequently \emph{memorize} parts of their training data~\cite{Feldman20,FeldmanZ20, ShokriSSS17}. This poses a threat when the model's training data contains privacy-sensitive information, as deployed models may regurgitate memorized private data~\cite{CarliniLEKS19, CarliniTWJHLRBSEOR21, somepalli2023diffusion}.

Differential privacy (DP)~\cite{DworkMNS06} offers a formal solution to this problem. Informally, training a model with (user-level) differential privacy offers a \emph{guarantee} that the model will not depend too heavily on the sensitive data contributed by any one individual.
Among other threats, this protects against the model \emph{memorizing} training data.
But current approaches to differentially private learning scale poorly, and greatly sacrifice the model's useful generalization capabilities in order to provably prevent memorization.

To address this issue, a growing line of work suggests to augment differentially private learning algorithms with access to \emph{public data}~\cite{AbadiCGMMTZ16, PapernotCSTE19, TramerB21, YuNBGIKKLMWYZ22, LiTLH22, AroraR22, DeBHSB22, MehtaTKC23, KurakinSCGTT22, PandaTSMM22, nasr2023effectively, tang2024private}. 
The goal then is to first leverage large troves of non-privacy-sensitive data to learn generic features---independent from any data owner's private data---which can then be efficiently \emph{finetuned} with DP on sensitive data.

For example, suppose a company wishes to train a model on a corpus of chat messages from its end users. While the content of these messages is sensitive, the general structure of a chat message (i.e., syntax, grammar, etc.) is not sensitive. 
Thus, the company may wish to leverage a model that was pretrained on a large public corpus of text (preferably including chat conversations) and then finetune this model on the specific sensitive content of the end users' messages.

Even in the the absence of any privacy concerns this \emph{pretraining} and \emph{transfer learning} approach has become the de-facto strategy for achieving state-of-the-art results across a variety of challenging tasks in computer vision and natural language processing.
Here, a generic ``foundation model''~\cite{Bommasani++21} is first pretrained on massive and weakly curated data---typically scraped from the public 
%(and thus non-sensitive) 
Internet.
Thereafter, the model can be efficiently finetuned on various downstream tasks~\cite{RadfordWCLAS19}. Generative models such as large language models even exhibit powerful \emph{in-context learning} abilities, where the model ``learns'' new tasks at inference time solely on the basis of a small number of examples~\cite{BrownMRSKDNSSAAHKHCRZWWHCSLGCCBMRSA20}.

The impressive performance of foundation models %on the basis of their large Web-scraped pretraining data 
naturally places these models as ideal candidates for private learning. Indeed, as the pretraining data comes from publicly available sources, the pretrained model is fully independent of individuals' privacy-sensitive target data. And since these models learn new tasks extremely (sample)-efficiently, they should be able to also learn these tasks \emph{privately} with only a minor impact on performance.
Thus, it is unsurprising that a growing body of work investigates the benefits of using Web-scale pretraining for private learning~\cite{LiTLH22, YuNBGIKKLMWYZ22, AroraR22, DeBHSB22, MehtaTKC23}, and showcases significant improvements in performance on canonical private learning benchmarks.

For example, on the ImageNet dataset~\cite{DengDSLLF09}, in the absence of any pretraining, the approach of~\citet{SanderSS23} achieves a top-1 accuracy of 39.2\%, under a fairly weak provable DP guarantee of $\epsilon=8$; more recent work of~\citet{TangPSM23} improved this slightly to the current state of the art: 39.39\%. This represents an almost $6\times$ increase in error rate compared to the best non-private model trained solely on ImageNet (at least $86.7\%$ accuracy)~\cite{TuTZYMBL22}. In contrast, when leveraging a dataset of 4 billion Web images for public pretraining, \citet{BerradaDSHSSKSB23} achieve an accuracy of $86.8\%$ at a much more reasonable privacy budget of $\epsilon=1$ (with comparable results obtained by~\citet{DeBHSB22,MehtaTKC23}).

In a similar vein, \citet{LiTLH22} and \citet{YuNBGIKKLMWYZ22} finetune large pretrained language models such as RoBERTa~\cite{LiuOGDJCLLZS19} and GPT-2~\cite{RadfordWCLAS19} to achieve strong performance on downstream tasks with differential privacy.
\citet{AroraR22} further argue that LLMs can be personalized to each individual's personal and sensitive data while incurring \emph{no privacy cost} (i.e., $\epsilon=0$) by leveraging the pretrained model's ``zero-shot'' abilities.
This line of work suggests we are getting close to ``solving'' private learning. Indeed, as Web-scraped datasets grow larger and larger, the ability of pretrained models to privately adapt (``for free'') to new tasks will only get better.

\smallskip
\noindent \textbf{This position paper challenges this view, and critiques the \emph{public-pretraining and private-finetuning} paradigm.}
We raise two (orthogonal) concerns that models trained in this manner may fail to be a) \emph
{private}, or b) \emph{useful}.
We thus question the validity of current findings in this area for informing real-world deployments of differential privacy.

Our primary criticism challenges the notion that pretraining on publicly-available Web data should be viewed as neutral (i.e., non-sensitive) from the perspective of user privacy:

\begin{quote}
    \emph{Pretraining data scraped from the Web may be sensitive itself; because a ``privacy-preserving'' finetuned model can still memorize its pretraining data, this causes direct harm and dilutes the meaning of ``private learning''.}
\end{quote}

Specifically, our critique raises issue with the privacy semantics when finetuning data is sensitive, but the pretraining data is considered to be public: as we explore, the latter assumption is mismatched with norms and expectations of what it colloquially means for a model to be private.

Beyond this core concern with the appropriateness of using publicly available data for privacy-preserving learning, we further posit that this paradigm might not be as useful as existing research suggests, and that it could even lead to a net \emph{loss} of privacy at training or deployment time:

\begin{itemize}
    \item Current private learning benchmarks likely overestimate the value of public pretraining by fixating on settings with highly overlapping public and private data distributions.
    \item Public pretraining performs best with massively large models that cannot be run on end-user devices, thereby trading off one form of privacy (DP for the sensitive finetuning data) for another (the model's users have to outsource private data to a third party).
\end{itemize}

Each of the three issues we raise are largely orthogonal to one another, and solving any one of them need not affect the others. For example, even if we were to develop benchmarks that accurately reflect private workloads, the core issue of the potential sensitivity of pretraining data would remain.

This work is a \emph{position paper}, which takes a critical view of the current state of the field and highlights several aspects we find problematic.
We thus put forward a call for solutions from the community -- while we offer some broad suggestions on potential ways to address our concerns, we (intentionally) stop short of technically exploring solutions, as each of these challenges deserves significant attention beyond the scope of this article.

%The two issues articulated above are largely orthogonal to our core concern about the purported ``privacy'' of pretraining on publicly accessible data. That is, even if these issues were fixed (i.e., by developing benchmarks that accurately reflect private workloads, and enabling client-side deployment of pretrained models), the core issue of the potential sensitivity of pretraining data would remain. Conversely, if we could pretrain models solely on non-sensitive data, secondary concerns about the utility of these models, and the privacy of their deployment, would remain.

\subsection{Paper Overview}

In the reminder of this introduction, we provide a broad overview of all three issues above, and outline some open questions and paths forward for the field.
For the interested reader, Sections~\ref{sec:public_sensitive},~\ref{sec:benchmarks} and~\ref{sec:outsource} then delve into more details to support our main arguments.
Finally, Section~\ref{sec:conclusion} provides some concluding remarks and future outlooks.

\smallskip \noindent
\textbf{1.~~The Web contains privacy-sensitive data.~~} 
Training data scraped from the Web is indeed \emph{publicly accessible}, but this does not imply that using this data in machine learning applications poses no privacy risks.

Individuals may put some data on the Internet with a specific context-of-use in mind.
For example, someone may post their contact information along with a research publication with the intent that it is used to contact that person about details of the publication.
Sensitive data about individuals could also be uploaded to the Internet unintentionally (or by third parties privy to this information). As a result, people often underestimate how much information about them is accessible on the Web~\cite{wsj22privacy}, and might not consent to their ``publicly accessible'' personal data being used for training machine learning models. 
%While emerging ``right-to-be-forgotten'' laws may aid those affected in deleting their online data, this procedure becomes more challenging when the data enters the training sets of a machine learning model.

%\npc{We're missing here a discussion of the fact that some things are not uploaded to the internet intentionally. The gamergate IRC logs are probably not intended to be used for GPT2 training data. Mention briefly here, and maybe forward reference to the discussion at length below.}

The question then is whether finetuning (with DP) on top of such publicly pretrained models should really be publicized as \emph{privacy-preserving}.  It is entirely possible that models might memorize a large fraction of their (public, yet sensitive) pretraining dataset.
Then if the model ever leaks this pretraining data, this still harms the privacy of the data subjects. Such situations could run the risk of eroding affected individuals' trust in differential privacy to appropriately protect their data in other settings (e.g., for collecting census data~\cite{AbowdACGHHJKLMMSSZ22}).

The guarantees of differential privacy are complicated enough to understand when no public data is involved~\cite{CummingsKR21}, even for researchers~\cite{McSherry16a, McSherry16b}. 
Asking data owners to distinguish between their ``public'' and ``private'' data further muddies the waters---especially because this distinction may not always be evident to data owners, or even knowable for some data.

Going forward, we argue that researchers should make the privacy ramifications of using certain public data sources clearer. Indeed, not all public data is created equal. Some public sources (e.g., Wikipedia) are highly curated and may pose low risks of containing sensitive information. Alternatively, some data sources might consist of public data that carries explicit consent to be used. It is an important open research question to understand if pretraining models solely on such stringently curated data can provide similar benefits for downstream tasks, while mitigating  privacy risks.

\smallskip \noindent
\textbf{2.~~Benchmarks conflate private and public distributions.} 
Even if we were to solve the core privacy issue above---for example by %either by agreeing that all publicly accessible data is fair to use for a ``privacy-preserving'' model, or by 
pretraining very powerful models solely on non-sensitive datasets---it remains unclear if these models will actually be \emph{useful} for privacy-sensitive downstream tasks.

We argue that the usefulness of the public-pretraining paradigm on private tasks is currently hard to assess, because existing benchmarks study ``private'' datasets that are not actually any more ``sensitive'' than the ``public'' dataset that is used for pretraining. In fact, the two are often drawn from the same (or from a similar) underlying distribution.

For example, when we transfer from ImageNet to CIFAR-10 (e.g., as by~\citet{TramerB21, DeBHSB22}),
\emph{every single} class contained in the CIFAR-10 dataset has an identical
class label in the ImageNet dataset!
So can we say that any ``private learning'' actually happened?
After all, training on ImageNet alone has already taught the model how to recognize a cat, or an airplane, or a dog---thus, (privately) finetuning on CIFAR-10 is merely performing a loose form of domain adaptation to classify low-resolution images more accurately.
Despite this critique, this is a standard evaluation metric for private ML with public data.

Of course, the aforementioned papers do not actually care about privately classifying CIFAR-10, \emph{per se}. Rather, they aim to provide and evaluate a general \emph{framework} for combining public and private datasets. %And since benchmarks that represent sensitive tasks are uncommon, the authors evaluate their approach on the same benchmarks that are commonly used to assess progress in non-private learning.
But as a result, we argue it is not clear that measuring progress of ``private'' learning on any of these benchmarks is at all meaningful. Specifically, are these benchmarks actually measuring progress in private learning? Or are they just a direct proxy for progress on non-private representation learning? 

The answer to this question likely depends on whether there exists an overlap between ``public'' and ``private'' data in \emph{real} privacy-sensitive applications.
We posit that this will not be the case in many applications, i.e., the privacy sensitive data to be finetuned on will come from a data distribution that is only poorly represented in the public pretraining data.
For example, machine learning on medical data is a canonical motivation for private ML, but the data distributions may not resemble those which are publicly accessible.

Unfortunately, it has already been shown that if the overlap between the pretraining and target distributions is small, then current methods for large scale pretraining may be less effective. For example, in the challenging (but perfectly-privacy) zero-shot setting, %while pretraining on a large dataset of 4 billion Web images leads to very high accuracy on ImageNet, it leads to poor accuracy on MNIST (worse than a logistic regression trained on MNIST)~\cite{PhamDGLYLTL21}. The reason for this surprisingly low performance is simply that there are very few images consisting entirely of handwritten digits on the Internet.
%Another example is medical datasets~\cite{veeling2018rotation, bejnordi2017diagnostic}, which are more interesting from a privacy perspective.
%Again in a zero-shot setting, the authors of BASIC-L (a recent representation learning method) write~\cite{PhamDGLYLTL21}:
foundation models tend to perform poorly on medical tasks. To illustrate, the authors of BASIC-L (a representation learning method) write:

\begin{quote}
    \emph{PCam [a skin lesion dataset] is perhaps the most sensitive dataset where
BASIC-L performs poorly. For such an important task, the top-1 accuracy of  BASIC-L (59.6\%) 
%and CLIP (63.0\%) are 
[is] far below the bars for practical deployments, [...] %%We remark that PCam is a binary classification task,
%so 
%the accuracy of BASIC-L and CLIP are 
[and] just slightly above random guessing. 
[...] As our training data are weakly crawled and automatically curated
from the internet, without any emphasis on medical images, our BASIC-L model cannot learn enough to perform well on PCam. [...] despite the benefits of open-vocabulary image classification models, they are not ready to be
deployed to tasks that require in-domain expertise \cite{PhamDGKLYYCLWTL23}}
\end{quote}

There is evidence that some deficiencies of zero-shot learning in these settings may be overcome by (non-private) \emph{finetuning}~\cite{radford2021learning}. But this need not always be the case. For example, while large language models pretrained on Internet text achieve impressive performance on a variety of downstream tasks~\cite{BrownMRSKDNSSAAHKHCRZWWHCSLGCCBMRSA20}, they still achieve poor performance when finetuned on (potentially sensitive) tasks that are only weakly represented online.%---such as legal contract analysis~\cite{HendrycksBCB21,YinH22}.

%Yet, there is also evidence that pre-training on ImageNet has minimal transfer benefits when fine-tuning on medical imaging~\cite{RaghuZKB19}.
%Regardless, the efficacy of transfer to poorly-represented domains under differential privacy remains underexplored.

%
%Concretely, the authors of the BASIC-L model~\cite{PhamDGLYLTL21} (which was pretrained on a proprietary augmented version of JFT with 6 billion images and finetuned on various downstream tasks) state: 
%

Understanding the effectiveness and limitations of transfer learning has been a major research direction in the non-private machine learning community, particularly through the lens of distribution shifts~\cite{KohSMXZBHYPGLDSGEHBLKPLFL21}.
%However, since data for privacy-sensitive tasks are often less represented in public data, this question becomes even more important for settings of practical interest.
Yet, it is not always clear whether conclusions from the non-private setting are valid when we introduce explicit privacy constraints, as data that is very privacy sensitive may be particularly poorly represented in the distribution of publicly available pretraining data.

% \npc{Too much detail? Enough to know that they say it's bad.}
%Note that without pretraining, models trained directly on the PCam dataset can achieve 89.8\% accuracy~\cite{veeling2018rotation}.

We thus call for privacy researchers to consider (or create) new benchmarks that more closely match envisioned deployments of private learning. 
%In particular, we encourage researchers to draw a distinction between ``intra-domain transfer'' and ``cross-domain transfer'' scenarios. The former---exemplified by the ImageNet to CIFAR-10 task above---is widely represented in common benchmarks, while the latter is likely more representative of many privacy-sensitive scenarios (e.g., pretraining on Web data and finetuning on private medical data). 
Such benchmarks could for instance leverage sensitive datasets that were publicly released for research purposes, such as e.g., MIMIC~\cite{johnson2016mimic} or the dataset from the infamous Netflix Prize~\cite{BennettL07}.

\smallskip \noindent
\textbf{3.~~Large private models require trusting cloud services.}
When we train a model with DP, this guarantees that anyone who can access the trained model cannot learn much about any individual training sample.
However this is orthogonal to any confidentiality considerations about who sees the data during training and inference.

Ideally, when dealing with personal data (e.g., private chat messages), the sensitive data would not leave the individual's device.
This is usually possible: the differentially private training could be decentralized (e.g., as in Federated Learning~\cite{mcmahan2017communication}\footnote{We remind that federated learning, even \emph{with} differential privacy, comes with its own associated privacy risks~\cite{ZhuLH19,BoenischDSSSP23}}), and the trained model could be shipped to people's devices for inference.

Unfortunately, unlocking the full power of large-scale public pretraining currently requires drastically scaling model sizes. With current techniques, most foundation models are impossible to train or serve on end-user devices. For example, MobileBERT~\cite{SunYSLWZ20}---a language model optimized for on-device inference---has about 25M parameters; this is between two and four orders of magnitude smaller than state-of-the-art language models considered in recent works on private finetuning of language models~\cite{LiTLH22, AroraR22, YuNBGIKKLMWYZ22}.

We thus encourage researchers in private machine learning to also take into consideration the scale of these models, and their privacy implications. An important direction for future work is to develop techniques for \emph{distilling}~\cite{HintonVD15} large foundation models into smaller, more efficient models that are tuned for a specific (private) task.\footnote{Recently, subsequent to the original appearance of this paper, there has been significant interest in the development of powerful small models that facilitate on-device computation (e.g., \cite{Gunasekar++23,LiBEDGL23,Anil++23}).}
\\

\noindent\textbf{Are these issues not also present in \emph{non-private} learning?}
While our paper focuses on the shortcomings of large-scale public pretraining for \emph{private} workloads, 
many of our criticisms may seem to apply more broadly to any application of pretrained models.
However we believe that these issues are especially important in privacy-sensitive applications.

First, the act of labeling the whole Web as ``public'' for machine learning purposes is particularly egregious when these models are explicitly touted as ``privacy preserving'', as this dilutes the meaning of ``privacy'' and may downplay the benefits of other uses of privacy enhancing technologies.

Second, a large overlap between pretraining data and common benchmarks need not be a concern if the goal of the benchmark is to measure \emph{absolute progress on the considered task}, rather than \emph{progress of generic learning techniques}.
Many standard benchmarks are useful in the former sense (e.g., ImageNet measures the ability to classify 1000 types of every-day objects).
But since these tasks are not privacy relevant, their use as benchmarks in the privacy literature solely serves the latter goal above: to evaluate the progress of generic (private) learning \emph{techniques}. In this case, a large overlap between pretraining and finetuning tasks is problematic.
%Second, regarding the overlap between pretraining and finetuning distributions in common benchmarks, we remark that benchmarks can be valuable for two orthogonal reasons: (1) for measuring progress on the specific task captured by the benchmark (e.g., ImageNet measures the ability to classify 1000 types of every-day objects); and (2) for assessing progress in generic \emph{learning techniques}.
%For the former goal, a large overlap between pretraining data and benchmark data need not be a concern: public pretraining genuinely improves our models' ability to classify the types of objects contained in ImageNet. But the tasks considered in most existing benchmarks are not directly privacy-relevant. Thus, the use of these benchmarks in the existing privacy literature solely serves the latter goal above: to evaluate the progress in generic (private) learning \emph{techniques}. In this case, a large overlap between pretraining and finetuning tasks 

%Finally, the privacy risks we identify with remote deployments of very large pretrained models only apply when users' data is indeed privacy sensitive. For many non-sensitive applications, such a deployment may be perfectly adequate.

\bigskip
\noindent\textbf{In the remainder of this paper} we study each of the three challenges in more detail, provide further evidence for our claims, and discuss conclusions of our work.

\section{Is publicly accessible data public?}
\label{sec:public_sensitive}

When a model is reported to be ``trained with differential privacy,'' it should mean something.
And if a model is trained with DP from scratch, it does mean something
very precise: no data specific to any individual training record will be memorized by the final model.\footnote{Informally speaking, if the training procedure satisfies $(\epsilon, \delta)$-DP, then with probability at least $1-\delta$ the inclusion of one individual training record changes the probability of observing any outcome by at most a factor $e^\epsilon$. Most DP models are trained with R\'enyi DP~\cite{AbadiCGMMTZ16,Mironov17}, which is translated into $(\varepsilon,\delta)$-DP to give more interpretable guarantees.} %We remind that differential privacy protects against a wide range of privacy issues beyond just memorization.}
In the common pretrain-publicly-then-finetune-privately paradigm, the privacy semantics are slightly different. 
The finetuning dataset enjoys the privacy guarantees bestowed by DP, but there is \textbf{absolutely no privacy} afforded to data in the pretraining dataset.
Our main argument in this section is that these privacy semantics, while rigorous and precise, fall short of satisfying several privacy norms in the manner they are generally used.
As such, we consider it detrimental to label the resulting models as ``privacy-preserving,'' as their guarantees are at odds with how most individuals would interpret such a claim. 

%Yet, the common pretrain-publicly-then-finetune-privately scheme only guarantees privacy with respect to the \emph{finetuning dataset}, and
%\textbf{not the original dataset} that is used for pretraining.
%In fact, there is no guarantee at all for the privacy of the pretraining data.

The issue comes from the fact that such a ``privately-trained'' model
will still leak details of individuals whose data were contained in the pretraining dataset.
And if a data subject notices this and asks ``if this model is private, why was my data leaked?''
the only possible answer to give is that this data was not part of the dataset that was considered worth protecting. Such an explanation would likely not be very satisfactory, as individuals may still view some publicly accessible data as sensitive---especially if, as we make the case here, it was not intentionally made public.\footnote{We comment that, even \emph{without} any public data, and \emph{with} DP training correctly implemented, there are still risks that could lead to catastrophic privacy violations, generally with very low probability of occurrence. As one extreme example (which occurs with exceptionally low probability), consider a run of DPSGD in which all added noise happens to be negligibly small, and thus has similar privacy risks to an unprotected model. Such failures of DP are possible, albeit unlikely, and outside the scope of our paper.}

This issue could be mitigated by pretraining models solely on data that is entirely
non-sensitive. Alternatively, we could ask data owners to provide explicit consent for their data to be used for machine learning. But then we have to ensure that the resulting privacy risks are very clearly communicated first. For example, if some user application were to ask ``please share your data to help improve this product,'' users may expect that their data will be shared with the application developer, but \emph{not} potentially with all \emph{other} application users.

Unfortunately, the value in pretraining currently seems to arise mostly from the fact that we are able to train on massive uncurated datasets. As a consequence of the size of these datasets, much of the collected content will inevitably come from uncertain origins with no explicit user consent, and requesting consent becomes challenging.

Such issues arise even for extremely well-studied and strongly supervised datasets such as ImageNet. Despite its ubiquitous use, the dataset contains sensitive content of individuals (e.g., images of children, nudity, etc.)~\cite{imagenet_nudity}.
Furthermore, some datasets have even been completely retracted on privacy grounds:
TinyImages~\cite{torralba200880} is a dataset of 80M images scraped from the Web which was later subsampled to create the CIFAR-10 dataset~\cite{Krizhevsky09}. This dataset has since been deprecated due to the discovery of offensive and derogatory images~\cite{BirhaneP21}.
Larger-scale datasets used for natural language processing are possibly even more challenging to curate.
These datasets are often hundreds of gigabytes~\cite{gao2020pile} to terabytes~\cite{HoffmanBMBCRdLCHWCHNMvdDDGOSERVS22} in size,
gathered mostly by scraping the Internet for any available text data, with minimal content filtering or curation.

  While such datasets contain, by definition, only data that is \emph{public} (in the sense of ``publicly accessible'' on the Internet), their use in machine learning still presents significant privacy risks, as illustrated by the following two (real) examples.

\subsection{Two Motivating Examples}

\paragraph{\emph{Intentionally shared data, for use in a particular context:}}

Consider again the case of a company that trains a language model on the text messages of its end users. 
The company initializes their model with the publicly available GPT-2 model, and then finetunes it with DP on its own corpus of private chat messages. 
The company then deploys the model and promises users that this model is privacy preserving!

Peter W.~uses the model and types: ``\texttt{The phone number of Peter W.~is:}'' and the model auto-completes his correct phone number (and also helpfully supplies his fax number, physical address, and email address). Peter W.~claims this model violates his privacy. The company assures Peter that it does not, since their implementation satisfies a state-of-the-art $(\epsilon=0.1, \delta=10^{-12})$ level of differential privacy (with respect to the data used for finetuning).
%
%His privacy was actually compromised when he posted his phone number to his personal Tumblr page ten years ago. John Smith does not fully understand. No one even visits his Tumblr page anymore (sadly). And what does any of this have to do with COVID variants?
His privacy was actually compromised when he posted his phone number along side some technical documents in a report to the government several years ago. 

Peter might not be fully satisfied with such an answer.
We argue that many people might react like Peter if personal information about
them were ever output by a ``privacy preserving'' machine learning model.

In fact, the above example is not hypothetical: Peter W.~is a real person, and the GPT-2 language
model does know his phone number for exactly the reason described above~\cite{CarliniTWJHLRBSEOR21}.
And he is not alone: even state-of-the-art production models like ChatGPT know the phone numbers of many people who placed their phone number online for one purpose only for it to be used during model training \cite{nasr2023scalable}.
%
%While there are likely not many people whose personal information has been memorized by current models, the fact that there exist some individuals whose data has been leaked significantly weakens privacy arguments.

Similar issues could arise for other modalities, for example for photos that individuals post online. These may be posted for a particular purpose, e.g., on an individual's homepage for professional purposes, or on a social media site for sharing memories with friends.
This does not imply that the subjects consent to all possible downstream uses---as one extreme example, consider a generative model trained on publicly accessible photos from the Web, that is then abused for deepfake pornography. This constitutes a clear privacy violation~\cite{ParisD19}, regardless of whether the model's training data was public or not.

Privacy violations could also arise if machine learning models create new ways of searching and linking data that was posted online anonymously or pseudonymously. 
Attacks of this nature have always been possible (e.g., using existing image search engines), but cutting-edge advances in ML make (and will continue to make) them easier and easier to mount. 
For example, a model similar to CLIP~\cite{radford2021learning} that was trained on the entire Internet might enable problematic forms of image search (e.g., ``find all online images that match this photo'').

\paragraph{\emph{Not intentionally (or knowingly) shared data:}}
Not all content available on the Internet is posted intentionally.
In some cases, the original owner might not even know their information has been posted without their consent.

We begin with another example from GPT-2 where data was posted without the knowledge of the original author, and was then used to train a large language model.
The GPT-2 language model was trained on many webpages that were linked to by the social media website Reddit.
One of these articles was a transcript of an IRC conversation (of several thousand messages) between several individuals discussing sensitive political and societal topics.
These individuals were likely not aware their private conversations were recorded---let alone published for anyone to see.
However GPT-2 trained on this data as if it was intentional.

Another example involves surveillance camera footage. 
While surveillance cameras are ubiquitous, the public generally expects recordings to be available exclusively to security personnel. 
Nonetheless, many surveillance camera configurations employ minimal security, leading to livestreams of their feeds being publicly available~\cite{XuXC18}.

Data can also be published unintentionally.
Reportedly,\footnote{\url{https://www.theregister.com/2022/05/03/openai\_copilot\_cryptocurrency/}} in at least one example, a GitHub user unintentionally uploaded information about their cryptocurrency wallet to a public Git repository.
When Copilot (a coding assistant language model)~\cite{chen2021evaluating} was trained on this repository, it memorized this wallet's private key and allowed another user to withdraw money from the account.
In all likelihood, this user did not post the key to their cryptocurrency wallet on their GitHub intentionally.

In all of these examples, the data is publicly accessible on the Internet, but particular uses of this data constitutes a significant violation of privacy norms, particularly when used in combination with machine learning.
%For each of the examples above, 
One could argue that it is not the act of \emph{training} the machine learning model that caused a privacy leak---but rather the original act of \emph{publishing} this data on the Internet. 

\emph{We disagree!}
Machine learning models (and foundation models in particular) have the capacity to \emph{amplify} this leakage by disseminating this information in a much broader context, e.g., in new applications built on top of these pretrained models.
%As an analogy, ``doxing'' a victim by finding some (publicly accessible) sensitive information about them, and redistributing it broadly on social media, has the potential to cause significant harm~\cite{??}---even though this data was already publicly available.
As an analogy, a malicious person who \emph{doxes} another by ``releasing someone’s personal details onto the Internet in an easily accessible form'' is still causing harm even though ``these details may already be publicly available, but in difficult to access forms or distributed across various sources that obscure them from casual discovery'' \cite{douglas2016doxing}.\footnote{We emphasize that we are making a (subjective) \emph{moral} argument that the model trainer bears some culpability for propagating this information. Legally, their liability may differ based on jurisdiction, and we omit further discussion for simplicity.}

The recent work of~\citet{BrownLMST22} raises some related concerns, for the specific case of language models.\footnote{A concurrent work of \cite{Concurrent22} also highlights related concerns of ``correlation'' between datapoints in other modalities, including medical data.} Their core argument is that the privacy categorization of text data is inherently \emph{contextual} (see also, the more recent works of~\citet{MireshghallahKZTSSC24,HartmannSBETW23,NeelC23}). Thus, collecting text data from various contexts on the Web and aggregating it into a single public-facing language model may violate users' privacy expectations for this data. We expand on this argument here by: (1) considering other publicly available data modalities than just text; and (2) by discussing the potential erosion of trust in privacy technologies that could arise when conflating ``public'' and ``private'' data sources (see below).

%As an analogy, if you take someone's picture and phone number, print them on flyers and distribute them in every mailbox in your city, this represents an obvious privacy violation regardless of the fact that the picture existed on the Web and the phone number could be found in the local phonebook.
%\npc{Depending on how racy we want to get, a more compelling example would be either (1) doxing people on twitter by posting personal information---because htis information is now easier to get it's going to cause more harm even if it was avaialble otherwise, and even worse (2) if you find someone who posts, e.g., to OnlyFans, and then you recognize them, you shouldn't send those pictures to their friends and family even though those images were intentionally posted.}

\subsection{Privacy Expectations}

Given that not all content available online has the expectation of being used 
to pretrain large models (either because it was posted online for one
particular purpose, or because it was not even posted intentionally),
this raises reasonable privacy considerations for training on this ``public'' data.

By publicizing such models as being ``privacy preserving,'' the leakage of people's so-called ``public'' data could erode their trust in technologies such as differential privacy (which is not technically at fault here). This trust erosion could then also (mistakenly) carry over to settings where differential privacy \emph{is} applied properly to all collected data---e.g., the collection of census data in the US.

A counterargument might be that the issue here is merely one of properly \emph{educating} the public about differential privacy and its guarantees (in the face of public pretraining).
But it is already challenging for users to understand how the semantic guarantees of DP align with their own notions of ``privacy'', even when no pretraining is involved.
%to explain to people why, when you ask a (perfectly
%private) large language model what John Doe's email address is, why it
%might be able to guess the email address correctly (e.g., john.doe@gmail.com).
%
%The answer to privacy-experts is obvious: the model has generalized and knows
%that people often format their email address in one of just a few ways,
%and by chance it has guessed correctly here.
%
%To non-experts this is already challenging to understand.
%
By training non-privately we now compound this issue by introducing two ``tiers'' of data that are presumed to have very different privacy expectations attached to them. But if peoples' own privacy expectations do not match with this assumption, we run the risk that ``privacy-preserving'' models will cause real privacy harms.

\section{Are we still measuring progress on private learning?}
\label{sec:benchmarks}

%\ft{TODO: this needs editing}
%\npc{Tried to take a pass. I like this intro, maybe we should move some of it up above.}

The purpose of a benchmark is to measure progress on a particular task of interest.
The ImageNet benchmark, for example, measures the ability of classifiers to perform image classification across a range of everyday objects.

It is important to use the right benchmark---one where progress serves as an appropriate proxy for progress on the true task of interest.
For example, while researchers had historically used the MNIST dataset of hand-written digits to evaluate the performance of neural networks,
today this dataset is not seen as a reliable measure of progress.
This is both because it has become ``too easy'',
and also because lessons learned from squeezing the last $0.1\%$ test accuracy
out of the dataset often do not generalize to more interesting datasets~\citep[e.g.,][]{goodfellow2014explaining}.
%
%For instance, while Residual Neural Networks (ResNets)~\cite{HeZRS16} and Transformers~\cite{VaswaniSPUJGKP17,DosovitskiyBKWZUDMHGUH21} are the current state-of-the-art models for many vision domains (and beyond vision), the much simpler convolutional neural networks still dominate MNIST leaderboards.\footnote{\url{https://paperswithcode.com/sota/image-classification-on-mnist}}

We now make the case that current benchmarks used to evaluate privacy-preserving
machine learning are similarly insufficient,
and that we should instead study tasks that are more directly indicative of performance on real-world privacy-sensitive tasks.

\subsection{Intra-domain versus Cross-domain Finetuning}

Existing research that follows the public-pretraining-and-private-finetuning 
paradigm discussed earlier has so far focused mostly on the following tasks:

\begin{itemize}
\item Pretrain on CIFAR-100 or ImageNet and finetune on CIFAR-10~\cite{AbadiCGMMTZ16, PapernotCSTE19, TramerB21, PandaTSMM22}
\item Pretrain on Places365 and finetune on ImageNet~\cite{KurakinSCGTT22}
\item Pretrain on JFT or LAION-5B and finetune on ImageNet~\cite{DeBHSB22, MehtaTKC23}
\item Pretrain on text data scraped from the Web, and finetune on other text data scraped from the Web~\cite{YuZCYL21, LiTLH22, YuNBGIKKLMWYZ22}
\end{itemize}

% \ft{rephrased this below as it was a copy paste of the intro}
The issue is that many of the above settings have the property that the ``private'' finetuning data distribution is essentially a subset of the ``public'' data distribution.
For example, the data distribution from which CIFAR-10 is drawn is a \emph{strict subset} of the data distribution from which ImageNet is drawn. CIFAR-10 is drawn from (heavily downsampled) images from the Internet representing one of 10 objects: cats, horses, airplanes, etc. ImageNet is similarly drawn from images from the Internet representing one of 1000 objects, \emph{including each of the CIFAR-10 classes}.
So when we pretrain on ImageNet and privately finetune on CIFAR-10, is any ``private learning'' actually happening or are we merely performing a loose form of intra-domain transfer from high-resolution to low-resolution images?
The latter is a worthy goal, but the former is better aligned with what researchers try to understand in these settings: how to adapt to novel concepts which are only well-represented in sensitive data. 
We emphasize that the issue here is an overlap property of the data \emph{distributions}, which occurs even when the \emph{datasets} themselves are entirely disjoint.

The reliance on public pretraining with significant overlaps between ``public'' and ``private'' data distributions has become more prevalent in recent research.
\citet{AbadiCGMMTZ16} were, to our knowledge, the first to present private learning results with public pretraining. While their ``private'' dataset (CIFAR-10) and ``public'' dataset (CIFAR-100) share close similarities, the authors argued that ``the examples and the image classes [of CIFAR-100] are different from those of CIFAR-10.''
Follow-up papers then moved on to using larger pretraining sets (e.g, ImageNet~\cite{TramerB21, DeBHSB22}) but omitted the concern about class overlap between the private and public datasets.

The situation is analogous when benchmarking large language models. For example, the GPT-4~\cite{achiam2023gpt} and Gemini \cite{team2023gemini} papers present detailed analyses of how common evaluation benchmarks in NLP might overlap with the model's training data,
and the CLIP paper \cite{radford2021learning} analyzes how many ImageNet test images might be contained in their CLIP training dataset.

To highlight an extreme example of the questionable use of ``public pretraining,'' two recent works
in this area~\cite{DeBHSB22, MehtaTKC23} have pretrained on Google's JFT dataset~\cite{zhai2022scaling, sun2017revisiting} to achieve high accuracy (privately) on ImageNet.
However while ImageNet~\cite{DengDSLLF09} is a public dataset that any researcher is allowed to download, JFT is a proprietary dataset of 4 \emph{billion} Web images collected and labeled by Google that has not been made public.
Thus, in this setup the ``private'' dataset is actually more accessible than the ``public'' dataset!\footnote{The machine learning community has created large-scale, open-source datasets which can serve as an alternative to proprietary datasets like JFT.
Some notable such datasets include LAION-5B~\cite{SchuhmannBVGWCCKMWSKCSKJ22} and the Pile~\cite{gao2020pile}.
However, these particular datasets have been taken down, by the authors due to unintentional indexing of child sexual abuse material, and by a DMCA request, respectively. 
These developments leave nebulous the future of large open-source datasets, and raise further question about the nature of the contents of proprietary datasets.}
On top of this, the datasets are similar enough that parts of ImageNet are directly contained in JFT. The papers above do account for this by removing images from JFT that are near-duplicates of images from ImageNet. But the fact remains that here the private and public datasets are essentially identically distributed---and thus likely not representative of many real private learning scenarios.

Of course, the above papers merely adopted these benchmarks as illustrative examples of a public-to-private transfer setup.
Yet, we caution against such benchmarks becoming the standard for assessing progress in private learning \emph{techniques}.
This is because \textbf{these benchmarks make it hard to disentangle generic progress in unsupervised representation learning, from algorithmic improvements for private learning}. This issue is compounded by the fact that there is no consensus on what public pretraining data to use, and so different papers use a wide variety of incomparable public sources. For instance, prior work has presented results for private learning on CIFAR-10 while leveraging the following public data sources:
CIFAR-100~\cite{AbadiCGMMTZ16, asadian2022self};
\emph{unlabeled} CIFAR-100~\cite{asadian2022self};
ImageNet~\cite{DeBHSB22}; 
\emph{unlabeled} ImageNet~\cite{TramerB21}; 
2{,}000 random ImageNet samples~\cite{YuZCL21}; 
%\gnote{does this preceding paper~\cite{YuZCL21} use 2000 ImageNet samples for pretraining? I thought it's for gradient projection during private training. Which is still public data, but we haven't included any other results of this sort in the paper}
%\ft{yeah this is for gradient projection but I would argue it's the same thing essentially. I rephrased it slightly to be more general}
a \emph{single} $600 \times 225$ image engineered for pretraining~\cite{asadian2022self}, etc.

A potential explanation for some of these ``esoteric'' choices of pretraining datasets is that without any restriction on what can be considered as ``public'' data, some benchmarks (such as CIFAR-10) become uninteresting as the private task can essentially already be solved with \emph{perfect privacy}~\cite{AroraR22}.
%Foundation models like CLIP (and GPT-3) can solve image (and text)
%tasks without ever seeing data drawn specifically from any particular distribution.
%
%This is because modern foundation models like CLIP~\cite{radford2021learning} or GPT-3~\cite{BrownMRSKDNSSAAHKHCRZWWHCSLGCCBMRSA20} are no longer trained to explicitly solve any single task. Instead, they are trained on a generic task of ``next-token prediction'', and thereby attain the ability to freely \emph{generate} answers for many different tasks that may be covered in their training data. 
%Traditionally models have been trained to solve exactly one task:
%for example, a model trained on CIFAR-10 can (only) solve classify CIFAR-10
%images and not accurately classify data from other domains.
%
%Even models trained on the 1000-class ImageNet perform poorly when evaluated on
%the CIFAR-10 dataset---despite the fact that each of the 10 classes in CIFAR-10 (e.g., airplane, boat, car)
%are present in ImageNet.
%In contrast, modern foundation models like CLIP or GPT-3 are trained on a generic task of ``next-token prediction'', and thereby attain the ability to freely \emph{generate} answers for many different tasks that may be covered in their training data.
%As a result, the number of tasks that can be solved privately ``for free'' 
For example, OpenAI's pretrained CLIP model gets $96.2\%$ \emph{zero-shot} accuracy on the CIFAR-10 dataset (without any finetuning), just a few percentage points shy of the $\sim 99\%$ state-of-the-art using this dataset alone---despite the fact that CLIP never saw \emph{any} CIFAR-10 training data! Thus, by definition, CLIP achieves $96.2\%$ accuracy at $(\varepsilon,\delta)=(0,0)$-DP.
Similarly, \citet{PhamDGKLYYCLWTL23} achieve zero-shot 85.7\% top-1 accuracy on ImageNet, thus achieving perfect privacy and state-of-the-art accuracy even compared to models with much larger values of $\varepsilon$~\cite{MehtaTKC23,DeBHSB22}. 

Why do we believe this is a problem? 
After all, if it is possible to reach $96\%$ accuracy on CIFAR-10 without
even inspecting the training dataset, is this not actually private?
Again, our concern comes down to the fact that the underlying data distribution for both CIFAR-10 and CLIP's training dataset are the same: images scraped from the Internet.\footnote{Specifically, CIFAR-10 was collected as a subset of TinyImages, which itself is a dataset of 80 million images collected from the public Internet.
Similarly, CLIP's training dataset is also a dataset of 400 million images collected by downloading images from the public Internet.}

%And while in principle it is possible that there is an overlap between these two datasets, the authors perform extensive analysis and fail to reject (p > .05) the hypothesis that overlap TODO explains the gap.

As a result, unlike traditional forms of \emph{cross-domain transfer} where we must take
a classifier trained in one setting (e.g., pictures taken from the Internet) and transfer them to a completely new setting (e.g., classifying tumors), we have no such need here.
Instead, it is sufficient to adapt from one sampling from a distribution (images from the Internet) to another sampling of the same distribution---albeit a sample with slightly different preprocessing. Thus, we argue that benchmarks such as private learning on CIFAR-10 (with pretraining) are a bit like MNIST for general computer vision: there is little performance left to be squeezed out ($\approx 2\%$), and this marginal progress might not carry over to real privacy-sensitive settings where a close overlap between public and private data sources may not exist.

\subsection{Towards Better Benchmarks}

Ultimately, the question we want to answer is whether or not we (as a community) are making progress towards effective private learning. It is possible that we live in a world---which we will call \worldone---where all sensitive tasks we care about (e.g., medical classification tasks) are well represented by data that is publicly available, e.g., on the Internet. Alternatively, we might live in a different world---\worldtwo---where these sensitive tasks are \emph{not well} represented and mostly disjoint from any public data. 

Knowing which of these two worlds we are in is important! If we are in \worldone, then everything is great: for a sensitive task of interest, simply use a publicly pretrained foundation model and solve the task with either zero-shot learning~\cite{AroraR22} or minimal finetuning (barring the issues raised above about the pretraining data actually also being sensitive). In contrast, if we are in \worldtwo, then foundation models pre-trained on Internet data might be of little help in many privacy-sensitive settings, as exemplified by the minor benefit of transfer learning reported by~\citet{RaghuZKB19,PhamDGKLYYCLWTL23} on some medical tasks.

Crucially, we cannot know in which world we are in without \emph{collecting data that resembles privacy-sensitive tasks that we actually care about} (i.e., not CIFAR-10 or ImageNet).
We thus believe it is necessary for the private learning community to begin considering and curating new benchmarks---to properly disentangle advances in non-private representation learning from advances in privacy-preserving learning.
Such benchmarks could include existing sensitive datasets that have been released for research purposes, e.g., medical datasets~\cite{johnson2016mimic, IrvinRKYCCMHBSSMHSJLLPLN19, wang2017chestx, bejnordi2017diagnostic}, email corpora~\cite{klimt2004enron}, user reviews~\cite{BennettL07}, etc.

Of course, it is possible we do live in \worldone, and foundation models will perform well when tuned privately on sensitive tasks. This would be a very promising signal that private learning can be achieved in many real-world deployments. Nevertheless, concerns with the sensitive nature of pretraining sets (Section~\ref{sec:public_sensitive}), and the necessity to outsource large models (Section~\ref{sec:outsource}) could remain.

As a result, regardless of which world we are in, we encourage researchers to continue studying ways to make differentially private learning (without pretraining) better, as progress on this problem can also inform real-world deployments (even if these use some form of pretraining).

Inspired by this paper's first appearance, some subsequent works have studied the efficacy of cross-domain transfer in the differentially private setting.
We draw attention to the work of~\citet{BerradaDSHSSKSB23}, which performs public pretraining on large-scale public datasets such as ImageNet-21K and JFT, and private fine-tuning on a variety of datasets, including medical datasets like CheXpert~\cite{IrvinRKYCCMHBSSMHSJLLPLN19} and MIMIC-CXR~\cite{JohnsonPBGLDMH19}.
They show that this strategy is able to achieve reasonably high utility (i.e., close to the non-private SOTA) even for datasets belonging to such specialized domains, providing evidence towards being in \worldone.
We note that the paper lacks a baseline of private training from scratch, so it is tough to evaluate precisely how much the public pretraining helped.
As this work only offers results only for a few dataset pairs, it is important for the field to more broadly understand the efficacy of public pretraining for private ML before reaching any general conclusions.

\section{Large models require uploading private data}
\label{sec:outsource}

In order to achieve high accuracy, transfer learning with differential privacy currently requires using enormous pretrained models.
For example, state-of-the-art differentially private ImageNet models have over $250$ \textbf{million} parameters
and require over $100$ \textbf{billion} FLOPs to evaluate~\cite{DeBHSB22, MehtaTKC23}.
And while models of this size can still be run on high-end customer GPUs (250 million parameters require ``only'' 1 GB of memory), the size of state-of-the-art models is currently scaling at a much faster rate than (consumer) hardware.\footnote{For example, from 2019 to 2022 the size of the largest language models grew by a factor $100-1000\times$, while the transistor count and memory size of consumer GPUs grew by less than $2 \times$ (e.g., for Nvidia's GeForce series).}
(The largest language models, for example, are already larger than 500 \textbf{billion} parameters.)

As a result, using these enormous private models
will likely require that data owners upload their private data to some
remote cloud service.
This causes a direct tradeoff between the privacy of the individuals who provide the private training data and the privacy of the end users of the trained model.

To illustrate, suppose that the final model \emph{must} meet a minimum accuracy level to be viable (potentially at the cost of privacy). This accuracy could be reached in one of two ways: (1) use a very large pretrained model and finetune it with DP on sensitive data; (2) use a smaller model (possibly also pretrained) and finetune it without DP (or with very low privacy guarantees) on sensitive data.\footnote{Specifically, this hypothetical precludes finetuning a smaller model with strong DP guarantees, as the resulting accuracy would be too poor for use.}

The latter approach---using a smaller (non-privately-finetuned) model---has the advantage that the model can be evaluated locally
on each user's own device and thus poses no privacy risks to the model's ultimate end users. However, the sensitive training data is not guaranteed any protection from being memorized.
In contrast, the former approach of privately tuning a larger pretrained model has the advantage that we can achieve more stringent DP guarantees for the finetuning data without sacrificing model utility, but requires the model's end users to upload
their private data to a remote service.

The situation is actually slightly more complicated than this, because these models are also too large to be \emph{finetuned} locally (e.g., using federated learning).
While this burden could be reduced by employing parameter-efficient methods such as LoRA~\cite{HuSWALWC22}, enabling fine-tuning of the very largest models on-device seems far out of reach.
Furthermore, methods such as \emph{local} differential privacy~\citep{KasiviswanathanLNRS11} introduce too much noise to see practical deployment for such settings~\citep{BittauEMMRLRKTS17}.
As a result, owners of the sensitive training data face a tradeoff between either having their data be centrally collected for differentially private training, or keeping their data locally but not having any differential privacy guarantees.

%Aside from issues related to model size, it seems likely that organizations will continue to train and serve models remotely for a number of other reasons, including user convenience and to protect proprietary models. 
%Therefore, more broadly than related to concerns highlighted here, it seems that the privacy community will still have to grapple the with challenges surrounding remote computation with private data.

In principle, the confidentiality of outsourced sensitive data (both for training and inference) could be guaranteed using cryptographic techniques such as
fully homomorphic encryption~\cite{gentry2009fully, gilad2016cryptonets} or
secure multiparty computation~\cite{MicaliGW87, mohassel2017secureml}.
Unfortunately, for the time being we are several orders of magnitude away from being able to
efficiently apply these techniques in practice to large models.
Outsourcing ML workloads to trusted execution environments~\cite{ohrimenko2016oblivious, TramerB19} is another possible alternative, but the security (and scalability) of existing platforms are also currently limited and are unlikely to handle billion-parameter models.

\section{Where do we go from here?}
\label{sec:conclusion}

Public pretraining for private learning might not be the panacea that prior work has made it out to be, and we hope future work will carefully consider the use of public data when performing private training.
We conclude by outlining open questions and possible directions for future work:

\begin{itemize}
    \item \textbf{Articulate granular privacy considerations for Web data.}
    The private learning literature often falls back on a simplified dichotomy where all data is either ``public'' or ``private.''
    Yet, individual expectations about privacy are rarely so binary~\cite{nissenbaum2004privacy}. 
    
    We therefore encourage privacy researchers to advocate for a more responsible and granular approach to privacy when it comes to collecting training datasets---and especially datasets collected from the Internet.
    This could include developing techniques and procedures for establishing \emph{consent} for using Internet data as training data, for auditing existing datasets (including proprietarily collected ones) for sensitive content, and encouraging appropriate disclosure of 
    any privacy concerns (for example, in an accompanying datasheet~\cite{gebru2021datasheets}).
    This goal is part of a broader research direction focused on responsible dataset curation in machine learning~\cite{BenderGMS21,MitchellLLGMORTJK22}.
    
    \item \textbf{Construct privacy-friendly pretrained models.}
    It is an open problem whether one could train a (useful) foundation model that does not carry the burden of increased privacy risks. 
    %does protect the privacy of its training data. 
    One avenue could be to curate and pretrain models on large Internet datasets that do not contain any privacy-sensitive data.
    This would require careful consideration pertaining to which data should and should not be considered sensitive.\footnote{Consequently, we intentionally refrain from making any broad prescriptions on this front, as what is and is not private depends significantly on \emph{context}, with considerations including (but not limited to) the types of data, the application, and relevant privacy norms.} 
    Another approach would be to obtain explicit consent-of-use from data owners.
    Finally, it may be possible to pretrain a foundation model itself with differential privacy~\cite{AnilGGKM22,ponomareva2022training}. A core open problem here is how to set the right granularity for DP when training on data aggregated from various sources~\cite{BrownLMST22}. For example, while current works aim to pretrain language models with privacy at the level of individual \emph{sentences}~\cite{AnilGGKM22, ponomareva2022training}, such privacy guarantees are insufficient unless all references to a piece of sensitive information have first been rigorously eliminated or deduplicated~\cite{LeeINZECC22}.

    \item \textbf{Design better benchmarks to measure progress in private learning.}
    %Standard benchmarks such as ImageNet have played a crucial role in fostering progress in deep learning. But it is important to remind ourselves that progress on any specific benchmark only matters insofar as this progress generalizes to real tasks of interest. E.g., ImageNet serves as a proxy for the ability to classify a large number of real-world objects. And it turns out that large scale pretraining does help in improving performance not just on ImageNet, but also on the general task that the benchmark aims to reflect.
    Unfortunately, no good benchmark for \emph{private} learning currently exists.
    By this, we mean a benchmark that correlates well with the true task we care about, namely, efficient private learning in sensitive domains.
    Privacy-sensitive data is likely to have many characteristics that standard ML datasets lack. Thus, by re-purposing existing ML benchmarks for private learning we run the risk of promoting progress metrics that are only weakly correlated (or not at all correlated) with progress on real privacy-sensitive tasks. This issue is exacerbated with the avenue of public pretraining, since many canonical ML benchmarks can now be solved privately ``for free.'' %Yet, it is unclear whether this progress will translate to sensitive domains that are weakly represented among large public datasets (e.g., medicine, law, finance, etc.)
    We thus encourage the community to explore alternative benchmarks that more closely align with privacy-sensitive tasks of interest.\footnote{We believe that choosing the right benchmarks for private machine learning is a consequential task, deserving of significant exploration and justification.
    Such investigation is beyond the scope of the present position paper.
    As a result, we explicitly abstain from prescribing any specific benchmarks, and leave this for future work.}
    
    \item \textbf{Promote a holistic view on ML privacy.}
The issues discussed in this paper fall under the broader concern that the current ML privacy literature is predominantly too ``model centric.'' That is, most research focuses on the narrow (but important) problem of training a model (once) with DP. This line of research largely ignores broader privacy considerations around data collection (what data is collected, from what sources, and why?), data lifetimes (for how long is data kept, and how many models are trained on it?), model lifetimes, etc. We encourage further research on these important topics.
\end{itemize}

Finally, while the overall tone of our article is critical, we recognize and highlight that many recent works employing public data have played an important role in showing that differential privacy \emph{can} be preserved for certain complex machine learning problems, without suffering devastating impacts on utility. 
This is an important step forward for the field.
We focused our attention on what we believe to be some of the most important considerations in this area, in an effort to steer the community towards making the next important steps advancing private machine learning.

\section*{Acknowledgments}
We would like to thank Davis Blalock, Cynthia Dwork, Ivan Habernal, Tatsunori Hashimoto, Janardhan Kulkarni, Alexey Kurakin, Katherine Lee, Xuechen Li, Percy Liang, Ashwinee Panda, Thomas Steinke, Ivy Vecna, Sergey Yekhanin, and anonymous reviewers for their valuable comments on previous versions of this work.

GK is supported by an NSERC Discovery Grant, a Canada CIFAR AI Chair, and an unrestricted gift from Google.

\bibliographystyle{alpha}
\bibliography{biblio}
\end{document}